\documentclass[letterpaper, 10 pt, conference]{ieeeconf}  

\IEEEoverridecommandlockouts                              

\usepackage{graphicx}
\usepackage{colortbl}

\usepackage{times}
\usepackage{epsfig}
\usepackage{amsmath}
\usepackage{amssymb}
\usepackage{arydshln}
\usepackage{color}
\usepackage{caption}
\usepackage{url}
\usepackage{here}
\usepackage{cite}
\usepackage{comment}
\usepackage[ruled,vlined]{algorithm2e}

\renewcommand{\baselinestretch}{0.98}

\usepackage{multirow,array}
\newcommand{\argmin}{\mathop{\rm arg~min}\limits}

\title{\LARGE \bf
Spatio-Temporal Graph Localization Networks \\for Image-based Navigation
}

\author{Takahiro Niwa$^{1}$, Shun Taguchi$^{1}$, and Noriaki Hirose$^{1}$
\thanks{$^{1}$Takahiro Niwa, Shun Taguchi, and Noriaki Hirose are with Toyota Central R\&D Labs., INC, Japan
        {\tt\small niwa-takahiro@mosk.tytlabs.co.jp}}%
}

\begin{document}

\maketitle
\thispagestyle{empty}
\pagestyle{empty}

\begin{abstract}
Localization in topological maps is essential for image-based navigation using an RGB camera. Localization using only one camera can be challenging in medium-to-large-sized environments because similar-looking images are often observed repeatedly, especially in indoor environments. To overcome this issue, we propose a learning-based localization method that simultaneously utilizes the spatial consistency from topological maps and the temporal consistency from time-series images captured by the robot. Our method combines a convolutional neural network (CNN) to embed image features and a recurrent-type graph neural network to perform accurate localization. When training our model, it is difficult to obtain the ground truth pose of the robot when capturing images in real-world environments. Hence, we propose a sim2real transfer approach with semi-supervised learning that leverages simulator images with the ground truth pose in addition to real images. We evaluated our method quantitatively and qualitatively and compared it with several state-of-the-art baselines. The proposed method outperformed the baselines in environments where the map contained similar images. Moreover, we evaluated an image-based navigation system incorporating our localization method and confirmed that navigation accuracy significantly improved in the simulator and real environments when compared with the other baseline methods.
\end{abstract}

\section{\label{sec_intro}INTRODUCTION}
Autonomous mobile robots have been attracting attention because of their potential utility in daily applications such as transportation of objects, automatic cleaning, guidance, and patrols. As opposed to navigation systems using multiple LiDARs, some studies have tackled vision-based navigation using a monocular camera owing to their low cost, light weight, compact size, robustness, and high availability.

One of the existing techniques for visual navigation is visual simultaneous localization and mapping (SLAM), which simultaneously creates a map of the environment with a three-dimensional structure and estimates self-position~\cite{lsdslam,vslam,orbslam,orbslam2}. 
Although visual SLAM can produce a detailed map of the environment, it requires accurate camera calibration.
In addition, collision avoidance and robustness against environmental changes create challenges for visual navigation.

To address these issues, image-based navigation using topological maps has recently attracted significant attention~\cite{sptm,sptm_slam,sptm_ricoh,sptm_seannet,graphnav,chaplot2020neural,DFOX,deepmpc,hirose2020probabilistic}.
A topological map is a graph-structured map created from the image sequences obtained by the robot. 
Each node in the map contains a monocular camera image. 
Adjacent nodes are connected on edges based on image similarity~\cite{sptm} or reachability estimation ~\cite{DFOX}. 
During navigation, the robot identifies its own position as a node number on the topological map. The robot then generates subgoal images between the identified and destination nodes~\cite{dijkstra}. 
The navigation system derives velocity references to control the robot using the subgoal images and the current image from the robot~\cite{deepmpc,hirose2020probabilistic}. 
Further details are provided in the evaluation section of this paper.
\begin{figure}[t]
  \centering
  \includegraphics[width=0.99\hsize]{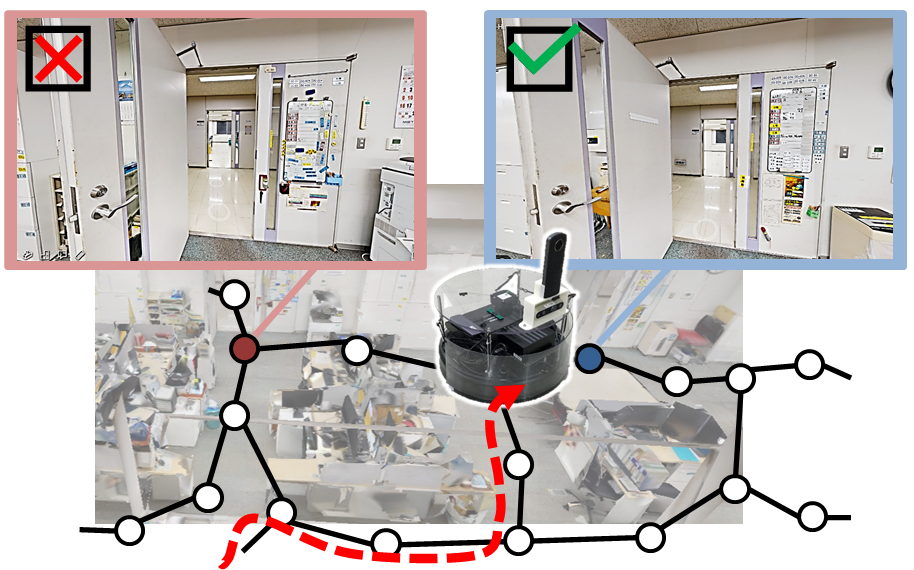}
  \caption{{\bf Localization for image-based navigation of robots, utilizing the graph structure of topological maps and time-series information from images observed by the robots.} Our method distinguishes between similar images via a recurrent-type graph neural network.}
  \label{fig_blockdiag_prior}
  \vspace{-5mm}
\end{figure}

In this study, we focus on localization in a topological map. Most localization methods for image-based navigation are based on image retrieval using the current robot image as the query, and the node images as the references \cite{sptm,sptm_slam,sptm_ricoh,sptm_seannet}. 
However, these methods pose two issues during navigation. 
Topological maps, especially in indoor environments, often contain similar images (as shown in Fig.~\ref{fig_blockdiag_prior}) because indoor environments such as office rooms, corridors, meeting spaces, and airports are often composed of repetitions of one environment. 
In such cases, baseline localization methods often select the wrong node, which leads to navigation failures.

Another issue is that collecting massive numbers of images and the ground truth~(GT) poses in a real environment is challenging. 
Hence, we cannot conduct supervised learning using the real images.

In this study, we propose a graph neural network-based localization method that can utilize spatial and temporal consistency using a topological map and time-series images from the robot (Fig.~\ref{fig_blockdiag_prior}). 
The proposed method is based on two intuitive ideas.
Adjacent node images in the map can help identify the correct nodes. Adjacent node images contain the same objects as in the current robot image and have similar appearances. This spatial consistency can be learned via a graph neural network using the proposed method. Another idea is to learn temporal consistency using robot time-series images via the LSTM layer. 
The history of the robot’s motion can eliminate the possibility of selecting similar but incorrect nodes.

In addition, e propose a semi-supervised learning method that utilizes simulator images with the GT pose in addition to real images without the GT pose.
The GT pose in the simulator images enables supervised learning to achieve accurate localization even in the real images~\cite{gonet,deepmpc}.

We evaluate the effectiveness of the proposed method by comparing it with several baseline methods. We also tested it on a robot navigation task using the Gibson simulator~\cite{igibson1,igibson2} and a real environment. The evaluation results show that the proposed method provides accurate localization and achieves highly accurate navigation in both simulated and real-world environments. The main contributions of this study are as follows:
\begin{itemize}
\item We proposed a novel graph neural network-based localization method. To the best of our knowledge, our study is the first application that uses a recurrent-type graph neural network for localization in a topological map.
\item We developed a semi-supervised learning method using real-world images without the GT pose and simulator images with the GT pose for sim2real transfer. 
\item We implemented an image-based navigation system that incorporated the proposed localization method into the evaluation.
\end{itemize}

\section{RELATED WORKS}
There is a long history of research on visual navigation for mobile robots.
We begin with a comprehensive discussion of visual navigation.
We then focus on visual localization, which is particularly relevant to our method.

\subsection{Visual Navigation System}
Visual navigation can be broadly divided into model- and learning-based approaches.
For model-based visual navigation, researchers have proposed solutions based on visual servoing and visual SLAM. 

Visual servoing \cite{rw1,rw2,rw3} controls an agent to minimize the difference between the current and goal states. 
Because the difference is defined in the image space, performance suffers when the environment changes or large obstacles occlude large parts of the environment.
Navigation methods based on visual SLAM \cite{lsdslam, vslam, orbslam, orbslam2} first use the camera images to construct a map that can be used by the robot to localize and compute actions to achieve a goal. The success of visual SLAM-based methods relies on acquiring an accurate metric model, and their performance decays when mapping failures occur.

To address these issues, multiple learning-based approaches have been recently proposed as image-based navigation. 
Image-based navigation using a topological map \cite{sptm, sptm_slam, sptm_ricoh, sptm_seannet, graphnav, chaplot2020neural} involves localization and planning from a topological representation of the environment that represents the connectivity between regions. 
The latest advances in reinforcement learning \cite{rw4,rw5,rw6,rw7} and imitation learning \cite{rw8,rw9,rw10} have also pushed the state of the art in image-based navigation.

\subsection{Visual Localization}
Visual localization using maps can be roughly divided into camera re-localization and visual place recognition.
Camera re-localization estimates the camera pose in Euclidean space in small known environments.
Several approaches such as direct camera pose regression \cite{posenet, relocnet}, coarse-to-fine \cite{camnet, sarlin2019coarse, revaud2019r2d2}, and structure-based approaches \cite{brachmann2017dsac, brachmann2018learning, brachmann2019expert, dsacstar} have been studied.
By contrast, visual place recognition is a task that involves retrieving images from a very large image database.
It is mainly based on hand-crafted or deep-learning-based image features \cite{arandjelovic2016netvlad, torii201524}.

Visual localization for image-based navigation retrieves the closest node on the graph, but does not estimate the camera pose in Euclidean space.
Therefore, it can be implemented for image retrieval on a graph, similar to visual place recognition.
In one of the earliest studies \cite{sptm}, localization was performed by estimating similarity using the Siamese network \cite{zagoruyko2015learning}.
This can be replaced by other image retrieval methods, such as NetVLAD \cite{arandjelovic2016netvlad}.

Image retrieval methods are estimated from only a single image; however, it is reasonable to use time-series observations for visual navigation. Our method employs a graph convolutional LSTM to improve the localization accuracy of a graph, which can handle time-series observations and spatial information from topological maps.

\section{PROBLEM STATEMENT}
We consider the problem of localizing a robot that moves along the edges of a topological map in an indoor environment. 
A topological map is a graph-structured map created from a sequence of images obtained by the robot during its past trajectories.
The images are held as nodes and spatially adjacent nodes are connected by edges in the topological map. 
Note that the self-position localized in this research is not the position in Euclidean space, but the index of the node closest to the robot.

In the following section, we define the topological map as a directed graph $G = (V, E)$, where $V=\{v_i\}_{i=1:n}$ is the set of $n$ nodes, and $v_i$ is the obtained image at node $i$. Also, $E=\{(r_k,s_k)\}_{k=1:m}$ is a tuple of $m$ edges, where edge $k$ is connected from source node $s_k$ to target node $r_k$.
The robot position is expressed as a sequence of node indices $Y=\{y_t\}_{t=1:T}$ corresponding to the observed images $O=\{o_t\}_{t=1:T}$.
Each robot position $y_t$ denotes the index of the node closest to the robot.
The topological localization problem is defined as follows:
{\it Given the topological map $G(V, E)$, find a current node $y_t$ at every time step $t$ using past observed images $O$.}

\section{\label{sec_method} PROPOSED METHOD}
As mentioned in the introduction, baseline localization methods have difficulty accurately localizing the self-position when multiple similar node images are contained in the topological map.
To solve this problem, we use the time-series images observed by the robot and spatial information from the topological map to consider spatial and temporal consistency for accurate localization.
In this section, we describe the proposed neural network configuration that utilizes spatio-temporal information sequentially.

\subsection{\label{overview} Spatio-Temporal Consistent Localization}

\begin{figure}[bt]
  \centering
  \includegraphics[width=0.99\hsize]{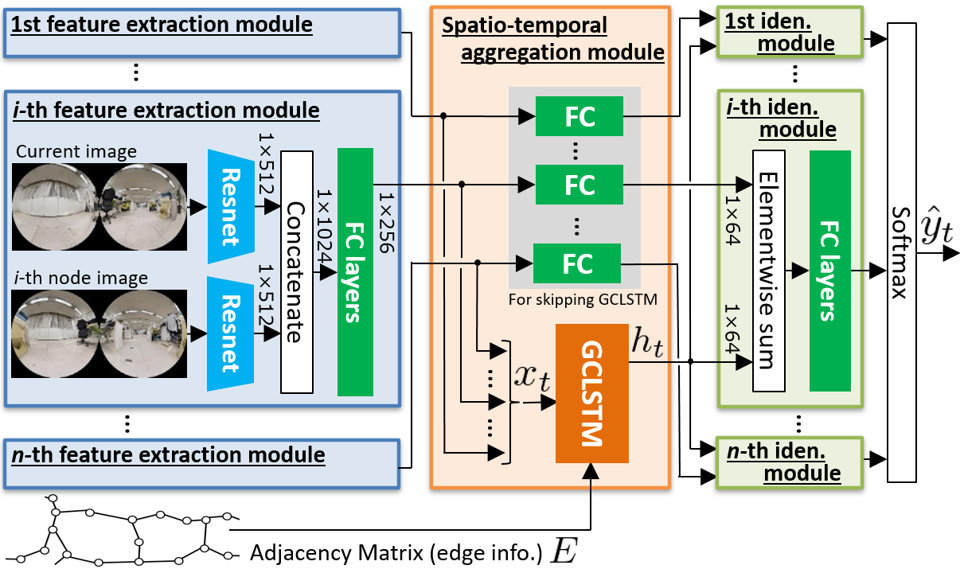}
  \vspace{0mm}
  \caption{{\bf Network structure of our proposed method with GCLSTM.} The Spatio-temporal aggregation module aggregates the features from the Feature extraction module to consider spatial and temporal consistency. Note that Feature extraction and Identification modules are performed for each node image.}
  \label{fig_gnn}
  \vspace{-5mm}
\end{figure}

We propose a novel graph neural network-based localization method that utilizes the spatial information from topological maps and time-series images observed by the robot. 
An overview of the proposed method is presented in Fig.~\ref{fig_gnn}.
Our model consists of three modules: a ``Feature extraction module'', a ``Spatio-temporal aggregation module'', and an ``Identification module''.
The Feature extraction module extracts a feature from the current node image and each remaining node image and calculates the relationship between them.
The Spatio-temporal aggregation module aggregates the features of neighboring and past nodes.
The Identification module calculates the localization probability for each node.
The details of each module are provided in the following.

\vspace{1mm}
\noindent
{\bf Feature extraction module}
This module consists of the ResNet encoder and fully connected~(FC) layers.
The current image $o_t$ and node images $V$ are encoded into feature vectors by the ResNet-18 encoder.
The ResNet encoder is used to extract the features of the images and reduce the dimensions. The FC layers are employed to extract the similarity between the current image and each node image. The features extracted from the FC layers are fed into the Spatio-temporal aggregation module. 

In inference tasks, ResNet-18 for the node images can be calculated offline to reduce the online computational load.

\vspace{1mm}
\noindent
{\bf Spatio-temporal aggregation module}
This module consists of graph convolutional LSTM~(GCLSTM)~\cite{gclstm} and a skip path with the FC layers.
We employ GCLSTM to simultaneously handle temporal information from time-series observations and spatial information from the topological maps.
GCLSTM is an extended model of a graph convolutional neural network~(GCN), which is a general and effective framework for learning representations of graph-structured data.
We introduced this graph-convolutional strategy into our topological localization.

The graph convolution function in GCN is generally computed as the weighted sum of the features corresponding to each node and its neighbor nodes.
By introducing this in GCLSTM, we can aggregate the features of the neighboring nodes of interest.
Moreover, GCLSTM can introduce time-series information to the graph convolution by using an LSTM-based structure.
Therefore, by using GCLSTM, we can introduce information, such as the node which closes to the node that has achieved high probability in previous observations is more confident, into the network.

While the GCLSTM layer outputs the overall features by convolving the features of neighboring nodes, it dilutes features of a self-node. 
To address this issue, we employ one FC layer to skip the GCLSTM layer to directly propagate the features of self-nodes to the later layers.
The FC layer reduces the dimensions of the features, which is equal to those of the GCLSTM outputs.
Note that the FC layers for each node share the same weights and biases.
The details of GCLSTM architecture are shown later.

\vspace{1mm}
\noindent
{\bf Identification module}
This module consists of FC layers and a softmax operator.
Two FC layers output the one-dimensional likelihood for each concatenated feature from the Spatio-temporal aggregation module. 
Finally, the likelihood of each node is fed into the softmax operator to estimate a localization probability.
The node with the highest probability can be estimated as the self-position.

\vspace{1mm}
In each module, we employ the ReLU activation function and batch normalization for all FC layers.

\subsection{\label{subsec_structure}GCLSTM Architecture}
Following \cite{gclstm}, the GCLSTM layer can be computed with the following equations:
\setlength{\arraycolsep}{0.0em}
\begin{eqnarray}
  \label{eq_gclstm1}
  i_t &{}={}& \sigma(\mathcal{G}_1(x_t, E) + \mathcal{G}_2(h_{t-1}, E) + w_{ci} \odot  c_{t-1} + b_i), \\
  \label{eq_gclstm2}
  f_t &{}={}& \sigma(\mathcal{G}_3 (x_t, E) + \mathcal{G}_4 (h_{t-1}, E) + w_{cf} \odot c_{t-1} + b_f), \\
  \label{eq_gclstm3}
  c_t &{}={}& f_t \odot c_{t-1} \nonumber \\
          && + \, i_t \odot \tanh(\mathcal{G}_5 (x_t, E) + \mathcal{G}_6 (h_{t-1}, E) + b_c), \\
  \label{eq_gclstm4}
  o &{}={}& \sigma(\mathcal{G}_7 (x_t, E) + \mathcal{G}_8 (h_{t-1}, E) + w_{co} \odot  c_{t } + b_o), \\
  \label{eq_gclstm5}
  h_t &{}={}& o \odot \tanh (c_t)
\end{eqnarray}
\setlength{\arraycolsep}{5pt}
where $x_t \in \mathbb{R}^{n \times d_x}$, $h_t \in [0,1]^{n \times d_h}$, and $c_t \in \mathbb{R}^{n \times d_h}$ are the input, cell output, and cell state, respectively.
Additionally, $\odot$ and $\sigma$ are the Hadamard product and sigmoid function, respectively, and $i,f,o\in [0,1]^{n \times d_h}$ are the input, forgetting, and output gates.
Weights
and biases $b_i,b_f,b_c,b_o\in \mathbb{R}^{d_h}$ are parameters of the model to be optimized in training.
In addition, $\mathcal{G}_{k,k=1:8}$ is an arbitrary graph convolution function that aggregates the features of neighboring nodes.
In this study, we used a graph isomorphism network (GIN) \cite{chebcomb}, which is known for its simple architecture and high discriminative/representational power.
The GIN updates the node representations as follows:
\begin{eqnarray}
  \label{eq_gin}
  \mathcal{G}_{k}(x_t, E) = {\rm MLP}_k((1+\epsilon_k)x_{t,i} + \Sigma_{j \in \{1:n\}}x_{t,j})
\end{eqnarray}
where ${\rm MLP}_k(\cdot)$, \, $\epsilon_k$, and $j$ are the multilayer perceptron consisting of two FC layers, weight coefficient, and indexes of neighboring nodes for node $i$, respectively.

The GCLSTM layer is structured to output $h_t$ by inheriting the past cell output $h_{t-1}$ and cell state $c_{t-1}$ in addition to the input $x_t$.
Because the cell output $h_{t-1}$ is the output of the prior step, it corresponds to short-term memory, and the cell state $c_{t-1}$ corresponds to long-term memory that is sequentially updated inside the cell according to (\ref{eq_gclstm3}).
In addition, because the input $x_t$ aggregates the information of the neighboring nodes using the graph convolution function, it is possible to infer, for example, that the current self-position is located around a node that showed a feature that is highly likely to be the self-position several steps in the past.
Therefore, by using the GCLSTM layer in the proposed method, it is possible to estimate the self-position using the features of the past and surrounding nodes in the long and short terms. 

\subsection{\label{sec:data}Training process}
To train the models shown in Fig.~\ref{fig_gnn}, we utilize the tuples $(O', G, Y)$ from the training dataset.
Here, $O'=\{o_{t}\}_{t=T-\tau:T}$ is the list of images observed by the robot from $T-\tau$ to $T$, and $G$ is the topological map. 
$Y=\{y_t\}_{t=T-\tau:T}$ is a list of GT indices for the nodes closest to each image in $O'$. In addition, $\tau$ is the step length for training.

We calculate $\hat{Y}$ by feeding $O'$ and $G$. Then, we optimize our model by minimizing the cross-entropy loss for $\tau$ steps from the estimated $\hat{Y}$ and the GT $Y$.

\subsection{\label{sec:semi}Semi-supervised Learning Method}
In the actual navigation, the robot cannot follow the exact image path from the topological map. The robot will deviate from its path depending on environmental changes, the robot's motor performance, and obstacle collision avoidance. To localize accurately in these cases, the time-series images for topological map $G$ and the time-series images for the robot's observation $O'$ for training should be obtained from different trials of teleoperation.
However, we cannot obtain $Y$ for the different time-series real images because there are no GT poses.

Hence, we propose a learning method that uses both simulated and real images to train our model to improve robot localization performance in the real world.
Because the simulator can provide the pose of each image $o_t$, the ground-truth node $y_t$ can be calculated for training. In this study, $y_t$ is computed as
\begin{eqnarray}
\label{eq_calcgt}
y_t = \argmin_{i=1:n} \{ \|p_t - p_i\| + \omega_m |\theta_t - \theta_i| \},
\end{eqnarray}
where $p_{\cdot} \in \mathbb{R}^{2}$ is the position in the $x-y$ coordinates [m] and $\theta_{\cdot} \in \mathbb{R}^{1}$ is the attitude angle in the yaw direction [deg].
Simulators also allow a large quantity of images to be collected in diverse environments.

If we use the simulated images only, it is expected that sim2real transfer issues will occur. 
Hence, we create $G$ and $O'$ from same time-series images for the real images. For the real images, we obtain $y_t$ from the node number of the nearest node in the time step. To achieve a sim2real transfer, we randomly mix the simulator dataset and the real dataset to create a mini-batch to improve the localization performance for the real images. Additionally, using the real dataset allows our model to learn static and dynamic environmental changes because our open dataset~\cite{gonet,lee2021largescale,taira2018inloc} includes pedestrians, changes in lighting conditions, and so on.

\subsection{\label{mapsampler} Map Sampler}
Because the memory of computational resources has an upper limit, when the number of nodes in the topological map $G$ is very large, it is impossible to train models by loading the images of all the nodes into memory. 
In this study, we devised and introduced a map sampler to construct a partial topological map $G^{'}$ with the number of nodes $n^{'}$ by sampling nodes from $G$ to include all elements of $Y$, the list of ground truths of the observed time-series images.

The method of constructing $G^{'}$ using the map sampler is as follows.
First, to include $Y$ in $G^{'}$, all the nodes in $Y$ are copied into an empty topological map $G^{'}$. 
Next, to add a new neighboring node to $G^{'}$, we randomly sample a node in $G$ that is connected by an edge with the node in $G^{'}$ and is not yet included in $G^{'}$, and then copy it to $G^{'}$.
We sample $G^{'}$ from $G$ by iterating this sampling process until the number of nodes in $G^{'}$ reaches the upper limit $n^{'}$. According to our map sampler, the sampled $G^{'}$ can always include nodes of $Y$ and can accommodate as many edges as possible to form a realistic map.

During training, the proposed method is trained by converting the tuple $(O^{'}, G, Y)$ into $(O^{'}, G^{'}, Y)$ using the map sampler described above.
The proposed map sampler also has an effect in terms of data augmentation because it randomly generates $G^{'}$ with different graph structures even when the same $G$ is used.

\section{\label{sec_eval}EXPERIMENTS}
We evaluated the proposed localization method for both localization and navigation tasks. First, we describe the dataset and the experimental setups. Subsequently, we present our results and compare them with several baselines.

\subsection{\label{subsec_data}Datasets}
In the training and testing, we used both real images without the GT pose and simulator images with the GT pose.
\subsubsection{Real images}
We employed the Go Stanford~(GS) Dataset~\cite{gonet,deepmpc}.
The GS dataset contains 360$^\circ$ camera images collected at the Stanford University campus. It contains 106,560 real images of 12 buildings and 39,307 simulator images of 36 environments from a mobile robot. Because the robot has no internal or external sensors to detect its global pose, the images in the GS dataset do not include the GT pose. 
Further details are shown in \cite{gonet,deepmpc}.

For the topological map using real image sequences, we created a node for every $m$ images in the sequences, and an edge was set from the previously created node to the newly created node. In this study, the value of $m$ was set at 7.

\subsubsection{Simulator images}
The simulator dataset was collected by the interactive Gibson simulator (iGibson)~\cite{igibson1,igibson2}. 
iGibson is a photorealistic robot simulator developed at the Stanford Vision and Learning Lab, which uses the Bullet physics engine to simulate interactions between objects. 

To obtain the virtual environments to be rendered by iGibson, we measured 50 rooms in our facility using the Matterport scanner 2. Then, we virtually teleoperated the robot to collect time-series images three times in each environment. The duration of one sequence was approximately 6 min. We separated the entire dataset into the following categories: 36 rooms with 108 trajectories for training, 7 rooms with 21 trajectories for validation, and 7 rooms with 21 trajectories for testing.

Observed images $O'$ and the topological map $G$ were created using two individual trajectories per environment. Hence, nine combinations (= 3 $O'$ $ \times $3 $G$ ) were prepared for training, validation, and testing.

A topological map of each room was created with nodes based on the GT pose of each node. 
Specifically, the first observation image in each trajectory is set as the first node $v_1$, and a new node $v_i \forall i \in \{2,...,n\}$ is added to the topological map when the position $p \, [\rm{m}] \in \mathbb{R}^{2}$ in the $xy$ coordinate and the attitude angle $\theta \, [^\circ] \in \mathbb{R}^{1}$ in the yaw direction satisfy the following equation: 
\begin{eqnarray}
  \label{eq_addnode}
  \|p - p_{i-1}\| + \omega_m |\theta - \theta_{i-1}| > \alpha_{th},
\end{eqnarray}
where $i$, $\omega_m$, and $\alpha_{th}$ are the index of the node in the topological map, the weight factor to balance the relative magnitude of the position and attitude, and the threshold for creating a new node, respectively.
In this study, we set $\omega_m$ = 0.025 and $\alpha_{th}$ = 1.0 by trial and error. 
The edge is set to be connected from $v_{i-1}$ to $v_{i}$.
In addition, to ensure closure of the map loop, the edge was connected from the most recently created node to node $v_j \forall j \in \{1,...,i-2\}$ when $\|p - p_{j}\| + \omega_m |\theta - \theta_{j}| > \alpha_{th}$ was satisfied.

\subsection{Experimental Setup}
We set the length of the time-series of data during training to $\tau$=90, and set the number of nodes in the map to be extracted by the map sampler described in section \ref{mapsampler} to $n^{'}$=200. 
For data augmentation, we randomly set the deviations in brightness, contrast, and saturation to $\pm$0.1 and hue to $\pm$0.05 for the set of robot-observed and node images. 
As the convolutional neural network that extracts image feature vectors, we used ResNet-18 \cite{resnet}, which was pre-trained with ImageNet \cite{imagenet}.

The learning rate of our model except for ResNet-18 was set to 0.001. We provided a smaller learning rate of 0.00001 for ResNet-18 to be fine-tuned.
The proposed method was trained until the minimum value of the validation loss was no longer updated for 1000 consecutive iterations.

\begin{table*}[tbh]
\centering
\caption{{\bf Performance comparison of localization with baseline methods in unseen environment.} Table shows the accuracy (AC) [\,\%\,], accuracy within one edge (AC*) [\,\%\,], pose error (PE) $\rm{[\, m+\omega_m\ast^\circ \,]}$ and map error (ME) [\,edge\,] in localization.}
\label{tbl_loc_result}
\begin{tabular}{l|cc|cc|cc|cc}
 & \multicolumn{2}{c|}{Not deviated} & \multicolumn{2}{c|}{Deviated $\leq 1.0$} & \multicolumn{2}{c|}{$1.0<$ Deviated $\leq 2.0$} & \multicolumn{2}{c}{Real image}                                                         \\
 Model & AC / AC* & PE / ME & AC / AC* & PE / ME   & AC / AC* & PE / ME & AC / AC* & PE / ME \\
  \hline \hline
  Pixel MSE \cite{4775883} & 0.751 / 0.840 & 1.463 / 1.987 & 0.632 / 0.758 & 1.945 / 2.462 & 0.284 / 0.472 & 3.300 / 4.561 & 0.736 / 0.820 & - / 28.423 \\
SSIM \cite{wang2004image}  & 0.837 / 0.920 & 0.668 / 0.958 & 0.758 / 0.874 & 0.902 / 1.114 & 0.290 / 0.472 & 3.183 / 4.294 & 0.744 / 0.830 & - / 22.259     \\
  SiameseNet \cite{sptm}  & 0.785 / 0.955 & 0.427 / 0.543 & 0.704 / 0.920 & 0.619 / 0.755 & 0.331 / 0.593 & 2.149 / 3.164 & 0.708 / 0.838 & - / 13.650 \\
  NetVLAD \cite{arandjelovic2016netvlad} & 0.788 / 0.929 & 0.568 / 0.847 & 0.725 / 0.891 & 0.777 / 1.127 & 0.328 / 0.493 & 2.824 / 4.190 & 0.760 / 0.856 & - / 12.603 \\
  \textbf{Our method} & \textbf{0.851} / \textbf{0.981} & \textbf{0.225} / \textbf{0.252} & \textbf{0.798} / \textbf{0.950} & \textbf{0.353} / \textbf{0.462} & \textbf{0.383} / 0.604 & \textbf{1.937} / \textbf{2.926} & \textbf{0.775} / \textbf{0.894} & - / \textbf{7.038} \\
  \quad w/o GCLSTM & 0.779 / 0.941 & 0.505 / 0.719 & 0.696 / 0.892 & 0.770 / 1.075 & 0.335 / 0.579 & 2.157 / 3.199 & 0.653 / 0.778 & - / 20.218 \\
  \quad w/o skip & 0.823 / 0.970 & 0.303 / 0.361 & 0.756 / 0.931 & 0.508 / 0.685 & 0.359 / 0.600 & 2.206 / 3.025 & 0.761 / 0.872 & - / 8.426 \\
  \quad w/o real image & 0.839 / 0.972 & 0.289 / 0.384 & 0.782 / 0.941 & 0.427 / 0.549 & 0.380 / \textbf{0.613} & 2.156 / 3.118 & 0.718 / 0.831 & - / 16.728 \\
\hline
\end{tabular}
\end{table*}
\subsection{Result: Localization}
The baseline methods used for comparison with the proposed method in terms of the localization performance are as follows.
\begin{enumerate}
    \item {\bf Pixel MSE \cite{4775883}:} The node with the smallest pixel-wise mean squared error~(MSE) in each channel is localized as the current node.
    \item {\bf SSIM \cite{wang2004image}:} Structural similarity index measure~(SSIM) is used to measure the similarity between two images, considering the distribution of pixel values, contrast, and structure. The node image with the highest similarity to the observation is localized to the current node. 
    \item {\bf SiameseNet \cite{sptm}:} SiameseNet \cite{sptm} estimates the similarity between two input images. The node image with the highest similarity to the observation is localized as the current node. We trained this model in the same manner as in \cite{sptm} and with the same dataset as our method.
    \item {\bf NetVLAD \cite{arandjelovic2016netvlad}:} We employed the pretrained model (VGG-16 + NetVLAD + whitening, trained on Pittsburgh dataset) for better results. This is a convolutional neural network (CNN) architecture that is directly trainable in an end-to-end manner directly for place recognition tasks.
\end{enumerate}

For the ablation study, we evaluated our method, and our method excluding the GCLSTM layer, the skip path with the FC layer, and real images from the training dataset (semi-supervised learning methodology described in section \ref{sec:semi}). 
Our method excluding the GCLSTM layer consists of four FC layers with ReLU and batch normalization. 

Table \ref{tbl_loc_result} presents the results of the numerical experiments on the unseen (test) datasets.
The table shows the accuracy (AC) [\,\%\,], accuracy within one edge (AC*) [\,\%\,], pose error (PE) $\rm{[\, m+\omega_m\ast^\circ \,]}$, and map error (ME) [\,edge\,] for the four data categories.
The category "Not deviated" refers to the simulator dataset in which the time-series images and the topological maps are created from the same trajectory.
"Deviated$\leq1.0$" and "$1.0<$Deviated$\leq2.0$" refer to a simulator dataset of time-series images and topological maps from different trajectories; the distance $\|p_t - p_{y_t^*}\| + \omega_m |\theta_t - \theta_{y_t^*}|$ between the observed image $o_t$ and the GT node $y_t^*$ is less than 1.0 in the former, and greater than 1.0 and less than or equal to 2.0 in the latter.
"Real image" refers to a dataset consisting of real images (GS dataset \cite{gonet,deepmpc}).
The numbers in the table represents the average of the localization results. 

As can be observed from Table \ref{tbl_loc_result}, the proposed method outperformed all baseline methods on all metrics. 
For the ablation study, our method was slightly inferior to our method without real images when testing with data in "$1.0<$Deviated$\leq2.0$", but showed high performance in the other metrics. 
It is considered that this occurred because our method without real images was trained only on the simulator images and optimized for them.

Moreover, the map error in the real data is relatively large for all methods. Because the topological map with the real images does not have the loop-closed points owing to the lack of GT poses, all methods estimate almost the same node at close poses, albeit with large map errors in some cases.
Note that the time-series images of the real image do not deviate from the topological maps. 
In Section \ref{subsec_nav}, we evaluate the navigation performance of our method in largely deviated scenes in real environments.

\begin{figure*}[t]
  \vspace{1.5mm}
  \begin{center}
    \includegraphics[width=0.99\hsize]{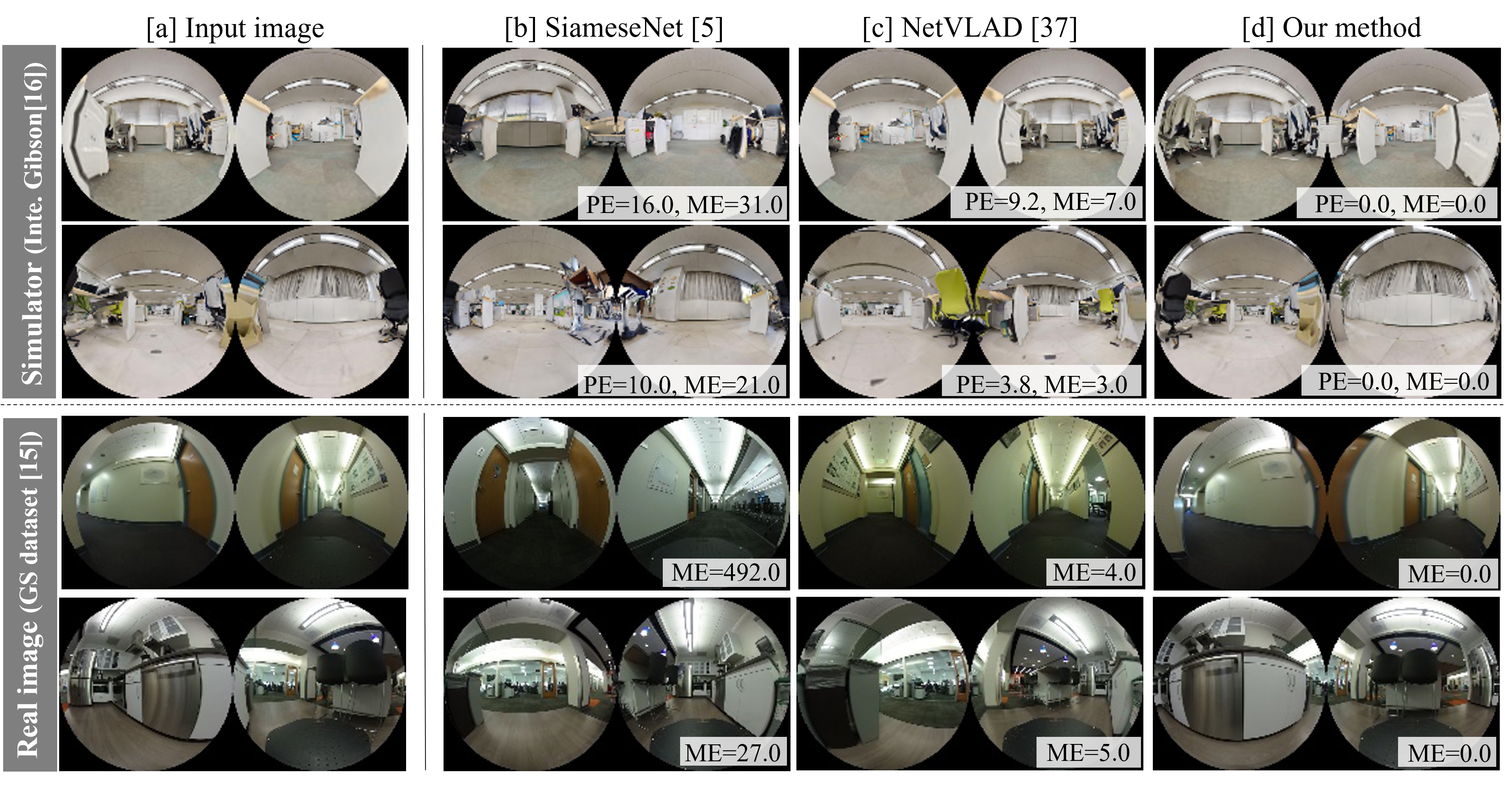}
  \end{center}
  \caption{{\bf Excerpts of node images localized by the baseline method (SiameseNet\cite{sptm}) and the proposed method (Ours) for the robot's observed image.} The baseline method misestimated a location where the images are similar, while the proposed method is more accurate in its estimation.}
  \label{tbl:qualiative}
  \vspace{-3mm}
\end{figure*}
Fig. \ref{tbl:qualiative} shows examples of node images localized by the baselines and proposed method for images observed by the robot.
From left to right, Fig. \ref{tbl:qualiative} shows the image observed by the robot and the localized node images produced by SiameseNet, NetVLAD, and our method, respectively.
The prediction error in pose $\rm{[\, m+\omega_m\ast^\circ \,]}$~(PE) and that in the edge distance of the map~(ME) [edge] are given at the bottom of the images.
As shown in Fig.~\ref{tbl:qualiative}, in some cases the baseline method significantly misestimated the self-position when the topological map contains multiple similar node images.
However, the proposed method localized the self-position accurately even when the topological map contained multiple similar node images.

\subsection{\label{subsec_nav}Result: Navigation}
\subsubsection{Overview of navigation system with our localization}
\begin{figure*}[t]
  \vspace{1.5mm}
  \begin{center}
      \includegraphics[width=0.95\hsize]{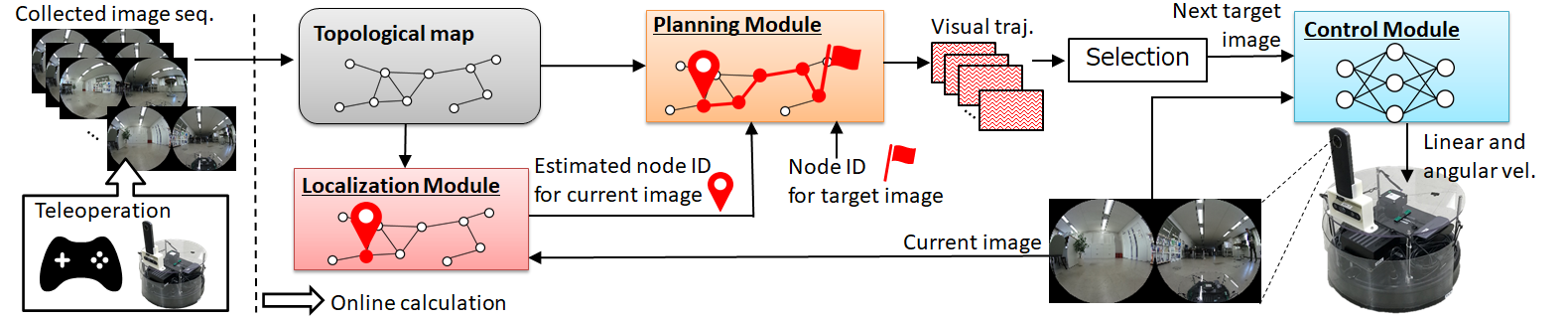}
  \end{center}
	\caption{\small {\bf Block diagram of image-based navigation system with our localization.} ``Localization module'' is our proposed localization with the GCLSTM. ``Planning module'' generates the visual trajectory from the current node to a target node. ``Control module'' derives linear and angular velocities to control the mobile robot.}
  \label{f:block_nav}
  \vspace{-3mm}
\end{figure*}

In addition to the sole evaluation of localization, we evaluate our proposed localization approach on the navigation system. Fig.~\ref{f:block_nav} shows a block diagram of the proposed navigation system with our localization. The following three modules are used: $\rm\hspace{.18em}i\hspace{.18em}$) localization, $\rm\hspace{.08em}ii\hspace{.08em}$) planning, and $\rm i\hspace{-.08em}i\hspace{-.08em}i$) control modules.
\begin{enumerate}
\renewcommand{\labelenumi}{\roman{enumi})}
    \item {\bf Localization module} estimates the number of nodes that correspond to the current robot position in the given topological map. We implemented our model in this module to evaluate our method on navigation.
    \item {\bf Planning module} generates subgoal images from the current node to the destination node. Dijkstra's method~\cite{dijkstra} was applied to minimize the number of images to shorten the navigation time. 
    \item {\bf Control module} derives the linear and angular velocity from the next subgoal image from ``selection'' and the current robot image. We provide the control policy of the DVMPC~\cite{deepmpc} to robustly control the mobile robot toward the subgoal position without collisions. 
\end{enumerate}
In our system, we calculate these modules every 3 fps until the robot arrives at the target position.

\begin{table}[tbp]
  \caption{{\bf Quantitative results for image-based navigation in unseen simulator environment}. The table shows the success rate (SR), collision rate (CR), time over rate (TR), and coverage rate (CovR) [\%].}
  \label{tbl:result}
  \centering
  \resizebox{1.0\columnwidth}{!}{
  \begin{tabular}{llcccc}
    \hline
    Env & Method  & SR & CR & TR & CovR \\
    \hline \hline
    \multirow{2}{*}{Area1}
            & SPTM~\cite{sptm} & 0.27 & 0.69 & 0.04 & 72.63 \\
            & SPTM with DVMPC~\cite{sptm,deepmpc} & 0.77 & 0.06 & 0.17 & 85.44 \\            
            & SPTM+ with DVMPC~\cite{sptm,deepmpc} & 0.78 & 0.10 & 0.12 & 84.74 \\
            & \textbf{Our method}                & \textbf{0.85} & \textbf{0.05}   & \textbf{0.01}   & \textbf{89.49} \\
    \hline
    \multirow{2}{*}{Area2}
            & SPTM~\cite{sptm} & 0.44 & \textbf{0.55} & 0.01 & 78.18 \\
            & SPTM with DVMPC~\cite{sptm,deepmpc} & 0.95 & \textbf{0.04} & 0.01 & 95.80 \\            
            & SPTM+ with DVMPC~\cite{sptm,deepmpc} & 0.91 & 0.06 & 0.03 & 94.11 \\
            & \textbf{Our method}                & \textbf{0.96} & \textbf{0.04}   & \textbf{0.00}   & \textbf{96.75} \\
    \hline
    \multirow{2}{*}{Area3}
            & SPTM~\cite{sptm} & 0.49 & 0.50 & 0.01 & 83.90 \\
            & SPTM with DVMPC~\cite{sptm,deepmpc} & 0.78 & 0.12 & 0.10 & 87.96 \\            
            & SPTM+ with DVMPC~\cite{sptm,deepmpc} & 0.80 & 0.17 & \textbf{0.03} & 86.35 \\
            & \textbf{Our method}                & \textbf{0.84} & \textbf{0.10}   & 0.06   & \textbf{88.10} \\
    \hline \hline
    \multirow{2}{*}{Mean}
            & SPTM~\cite{sptm} & 0.40 & 0.58 & 0.02 & 78.24 \\
            & SPTM with DVMPC~\cite{sptm,deepmpc} & 0.83 & 0.07 & 0.09 & 89.73 \\            
            & SPTM+ with DVMPC~\cite{sptm,deepmpc} & 0.83 & 0.11 & \textbf{0.06} & 88.40 \\
            & \textbf{Our method}                & \textbf{0.88} & \textbf{0.06}   & \textbf{0.06}   & \textbf{91.44} \\
    \hline
  \end{tabular}
  }
\end{table}
\subsubsection{Comparison to baselines in simulation}
First, we compared the navigation performance of our method with the following three baseline methods in a simulation.
\begin{enumerate}
\renewcommand{\labelenumi}{\roman{enumi})}
    \item {\bf SPTM \cite{sptm}:} We construct the same navigation system as \cite{sptm} and train their models with the same dataset as our method. Following \cite{sptm}, the localization is based on the SiameseNet.
    \item {\bf SPTM with DVMPC \cite{sptm,deepmpc}:} We replace the control module of SPTM with DVMPC~\cite{deepmpc}. 
    \item {\bf SPTM+ with DVMPC \cite{sptm,deepmpc}}: We apply our localization method without GCLSTM instead of SiameseNet in SPTM with DVMPC. The only difference between the two methods is the localization method.
\end{enumerate}

We chose three simulation environments and performed 100 trials in each environment. 
The distance between the robot's initial position and the goal node is within 10 [m] and is randomly generated in each trial.
Before navigation, we collected time-series images by teleoperating the virtual robot and created the topological map in each environment, following section~\ref{subsec_data}. 
In each trial, we stopped the navigation when the robot collided with the obstacles and when the total navigation time exceeded the threshold. 

Table \ref{tbl:result} shows the mean of the four metrics.
SR denotes the success rate to arrive at the target final position, CR denotes the collision rate where the robot collides with obstacles, TR denotes the time over rate where the 180 [s] threshold is exceeded, and CovR denotes the coverage rate against the desired trajectories between the start and goal positions. 

From Table \ref{tbl:result}, we can confirm that our navigation system with our localization method outperformed all baseline methods in all three environments. The main advantage of our localization method lies in the difference in performance between our method and SPTM+ with DVMPC.   

\begin{table}[tbp]
  \caption{{\bf Navigation performance in unseen real environments}. The table shows the success rate (SR) and coverage rate  (CovR) [\%].}
  \label{tbl:result_real}
  \centering
  \begin{tabular}{llcc}
    \hline
    Env & Method  & SR & CovR \\
    \hline \hline
    \multirow{2}{*}{Env1}
            & SPTM with DVMPC~\cite{sptm,deepmpc} & 0.70 & 85.00 \\            
            & \textbf{Our method}                & \textbf{0.90} & \textbf{95.00} \\
    \hline
    \multirow{2}{*}{Env2}
            & SPTM with DVMPC~\cite{sptm,deepmpc} & 0.50 & 73.80 \\            
            & \textbf{Our method}                & \textbf{0.60}  & \textbf{83.00} \\
    \hline
    \multirow{2}{*}{Env3}
            & SPTM with DVMPC~\cite{sptm,deepmpc} & 0.60 & 88.00 \\            
            & \textbf{Our method}                & \textbf{0.70}  & \textbf{91.00} \\
    \hline \hline
    \multirow{2}{*}{Mean}
            & SPTM with DVMPC~\cite{sptm,deepmpc} & 0.60 & 82.26 \\            
            & \textbf{Our method}                & \textbf{0.73}  & \textbf{89.66} \\
    \hline
  \end{tabular}
\end{table}
\subsubsection{Experiments with physical robot}
We evaluated our method with a physical robot in a real-world environment.
Navigation experiments were conducted with a vizbot, a small robot platform. The mobile base of the vizbot is the Roomba from iRobot. We utilized an Nvidia Jetson Xavier as the control personal computer and a Ricoh Theta S as a 360-degree camera to implement our image-based navigation. 

We compared our method to the best baseline method in the previous section, SPTM with DVMPC~\cite{sptm,deepmpc}.
We chose three environments and performed 10 trials in each environment. 
Other conditions in the physical robot experiments were the same as those in the simulator experiments.

Table \ref{tbl:result_real} shows the mean of the success rate (SR) and coverage rate (CovR).
As in the simulator environment, our method showed a better success rate and coverage rate against the baseline method in the real world. 

\section{CONCLUSIONS}
We proposed a localization method utilizing recurrent-type graph neural networks that uses the spatial information of the environment and the temporal information from the robot's trajectory.
The proposed method was trained on simulator images with the GT pose as well as real images without the GT pose for sim2real transfers. The evaluation results show that the proposed method outperforms the baseline method in localization and navigation tasks. 
Because the proposed method uses the graph structure of the topological map and the time-series information of the robot's observation images for localization, it can achieve an accurate estimation even when the topological map contains multiple similar node images.

In the future, we are planning to improve the localization performance in the largely deviated scene from the topological map.
Even while avoiding large obstacles, the robot needs to precisely localize its own position for more robust navigation.


\section{ACKNOWLEDGMENT}
We thank Kazutoshi Sukigara, Kota Sato, Yuichiro Matsuda, and Yasuaki Tsurumi for measuring the 3D environments utilized for collecting training images with the GT pose and evaluating our method for navigation.
We thank Hideki Deguchi for the evaluation for navigation, and Satoshi Koide and Keisuke Kawano for the technical discussion.
We would like to thank Editage (www.editage.com) for English language editing.

\bibliographystyle{IEEEtran}
\vskip-\parskip
\begingroup
\bibliography{loc}

\begin{thebibliography}{10}
\providecommand{\url}[1]{#1}
\csname url@rmstyle\endcsname
\providecommand{\newblock}{\relax}
\providecommand{\bibinfo}[2]{#2}
\providecommand\BIBentrySTDinterwordspacing{\spaceskip=0pt\relax}
\providecommand\BIBentryALTinterwordstretchfactor{4}
\providecommand\BIBentryALTinterwordspacing{\spaceskip=\fontdimen2\font plus
\BIBentryALTinterwordstretchfactor\fontdimen3\font minus
  \fontdimen4\font\relax}
\providecommand\BIBforeignlanguage[2]{{%
\expandafter\ifx\csname l@#1\endcsname\relax
\typeout{** WARNING: IEEEtran.bst: No hyphenation pattern has been}%
\typeout{** loaded for the language `#1'. Using the pattern for}%
\typeout{** the default language instead.}%
\else
\language=\csname l@#1\endcsname
\fi
#2}}

\bibitem{lsdslam}
J.~Engel \emph{et~al.}, ``Lsd-slam: Large-scale direct monocular slam,'' in
  \emph{European conference on computer vision}.\hskip 1em plus 0.5em minus
  0.4em\relax Springer, 2014, pp. 834--849.

\bibitem{vslam}
J.~Fuentes-Pacheco \emph{et~al.}, ``Visual simultaneous localization and
  mapping: a survey,'' \emph{Artificial intelligence review}, vol.~43, no.~1,
  pp. 55--81, 2015.

\bibitem{orbslam}
R.~Mur-Artal \emph{et~al.}, ``Orb-slam: a versatile and accurate monocular slam
  system,'' \emph{IEEE transactions on robotics}, vol.~31, no.~5, pp.
  1147--1163, 2015.

\bibitem{orbslam2}
------, ``Orb-slam2: An open-source slam system for monocular, stereo, and
  rgb-d cameras,'' \emph{IEEE transactions on robotics}, vol.~33, no.~5, pp.
  1255--1262, 2017.

\bibitem{sptm}
N.~Savinov \emph{et~al.}, ``Semi-parametric topological memory for
  navigation,'' \emph{arXiv preprint arXiv:1803.00653}, 2018.

\bibitem{sptm_slam}
D.~S. Chaplot \emph{et~al.}, ``Neural topological slam for visual navigation,''
  in \emph{Proceedings of the IEEE/CVF Conference on Computer Vision and
  Pattern Recognition}, 2020, pp. 12\,875--12\,884.

\bibitem{sptm_ricoh}
A.~Taniguchi \emph{et~al.}, ``Pose invariant topological memory for visual
  navigation,'' in \emph{Proceedings of the IEEE/CVF International Conference
  on Computer Vision}, 2021, pp. 15\,384--15\,393.

\bibitem{sptm_seannet}
X.~Li \emph{et~al.}, ``Seannet: Semantic understanding network for localization
  under object dynamics,'' \emph{arXiv preprint arXiv:2110.02276}, 2021.

\bibitem{graphnav}
K.~Chen \emph{et~al.}, ``A behavioral approach to visual navigation with graph
  localization networks,'' \emph{arXiv preprint arXiv:1903.00445}, 2019.

\bibitem{chaplot2020neural}
D.~S. Chaplot \emph{et~al.}, ``Neural topological slam for visual navigation,''
  in \emph{Proceedings of the IEEE/CVF Conference on Computer Vision and
  Pattern Recognition}, 2020, pp. 12\,875--12\,884.

\bibitem{DFOX}
X.~Meng \emph{et~al.}, ``Scaling local control to large-scale topological
  navigation,'' in \emph{2020 IEEE International Conference on Robotics and
  Automation (ICRA)}.\hskip 1em plus 0.5em minus 0.4em\relax IEEE, 2020, pp.
  672--678.

\bibitem{deepmpc}
N.~Hirose \emph{et~al.}, ``Deep visual mpc-policy learning for navigation,''
  \emph{IEEE Robotics and Automation Letters}, vol.~4, no.~4, pp. 3184--3191,
  2019.

\bibitem{hirose2020probabilistic}
------, ``Probabilistic visual navigation with bidirectional image
  prediction,'' in \emph{2021 IEEE/RSJ International Conference on Intelligent
  Robots and Systems (IROS)}.\hskip 1em plus 0.5em minus 0.4em\relax IEEE,
  2020, pp. 1539--1546.

\bibitem{dijkstra}
E.~W. Dijkstra \emph{et~al.}, ``A note on two problems in connexion with
  graphs,'' \emph{Numerische mathematik}, vol.~1, no.~1, pp. 269--271, 1959.

\bibitem{gonet}
\url{https://cvgl.stanford.edu/gonet/dataset/}, (accessed Jan. 18, 2022).

\bibitem{igibson1}
F.~Xia \emph{et~al.}, ``Interactive gibson benchmark: A benchmark for
  interactive navigation in cluttered environments,'' \emph{IEEE Robotics and
  Automation Letters}, vol.~5, no.~2, pp. 713--720, 2020.

\bibitem{igibson2}
------, ``Gibson env v2: Embodied simulation environments for interactive
  navigation,'' \emph{Stanford University, Tech. Rep.}, 2019.

\bibitem{rw1}
F.~Chaumette and S.~Hutchinson, ``Visual servo control. i. basic approaches,''
  \emph{IEEE Robotics \& Automation Magazine}, vol.~13, no.~4, pp. 82--90,
  2006.

\bibitem{rw2}
------, ``Visual servo control. ii. advanced approaches [tutorial],''
  \emph{IEEE Robotics \& Automation Magazine}, vol.~14, no.~1, pp. 109--118,
  2007.

\bibitem{rw3}
F.~Codevilla \emph{et~al.}, ``End-to-end driving via conditional imitation
  learning,'' in \emph{2018 IEEE international conference on robotics and
  automation (ICRA)}.\hskip 1em plus 0.5em minus 0.4em\relax IEEE, 2018, pp.
  4693--4700.

\bibitem{rw4}
S.~Hutchinson \emph{et~al.}, ``A tutorial on visual servo control,'' \emph{IEEE
  transactions on robotics and automation}, vol.~12, no.~5, pp. 651--670, 1996.

\bibitem{rw5}
G.~Kahn \emph{et~al.}, ``Self-supervised deep reinforcement learning with
  generalized computation graphs for robot navigation,'' in \emph{2018 IEEE
  International Conference on Robotics and Automation (ICRA)}.\hskip 1em plus
  0.5em minus 0.4em\relax IEEE, 2018, pp. 5129--5136.

\bibitem{rw6}
K.~Kase \emph{et~al.}, ``Learning multiple sensorimotor units to complete
  compound tasks using an rnn with multiple attractors,'' in \emph{2019
  IEEE/RSJ International Conference on Intelligent Robots and Systems
  (IROS)}.\hskip 1em plus 0.5em minus 0.4em\relax IEEE, 2019, pp. 4244--4249.

\bibitem{rw7}
D.~Mishkin \emph{et~al.}, ``Benchmarking classic and learned navigation in
  complex 3d environments,'' \emph{arXiv preprint arXiv:1901.10915}, 2019.

\bibitem{rw8}
A.~Pokle \emph{et~al.}, ``Deep local trajectory replanning and control for
  robot navigation,'' in \emph{2019 International Conference on Robotics and
  Automation (ICRA)}.\hskip 1em plus 0.5em minus 0.4em\relax IEEE, 2019, pp.
  5815--5822.

\bibitem{rw9}
E.~Wijmans \emph{et~al.}, ``Dd-ppo: Learning near-perfect pointgoal navigators
  from 2.5 billion frames,'' \emph{arXiv preprint arXiv:1911.00357}, 2019.

\bibitem{rw10}
Y.~Zhu \emph{et~al.}, ``Target-driven visual navigation in indoor scenes using
  deep reinforcement learning,'' in \emph{2017 IEEE international conference on
  robotics and automation (ICRA)}.\hskip 1em plus 0.5em minus 0.4em\relax IEEE,
  2017, pp. 3357--3364.

\bibitem{posenet}
A.~Kendall \emph{et~al.}, ``Posenet: A convolutional network for real-time
  6-dof camera relocalization,'' in \emph{ICCV}, 2015, pp. 2938--2946.

\bibitem{relocnet}
V.~Balntas \emph{et~al.}, ``Relocnet: Continuous metric learning relocalisation
  using neural nets,'' in \emph{ECCV}, 2018, pp. 751--767.

\bibitem{camnet}
M.~Ding \emph{et~al.}, ``Camnet: Coarse-to-fine retrieval for camera
  re-localization,'' in \emph{ICCV}, 2019, pp. 2871--2880.

\bibitem{sarlin2019coarse}
P.-E. Sarlin \emph{et~al.}, ``From coarse to fine: Robust hierarchical
  localization at large scale,'' in \emph{CVPR}, 2019, pp. 12\,716--12\,725.

\bibitem{revaud2019r2d2}
J.~Revaud \emph{et~al.}, ``R2d2: repeatable and reliable detector and
  descriptor,'' \emph{NIPS}, 2019.

\bibitem{brachmann2017dsac}
E.~Brachmann \emph{et~al.}, ``Dsac-differentiable ransac for camera
  localization,'' in \emph{CVPR}, 2017, pp. 6684--6692.

\bibitem{brachmann2018learning}
E.~Brachmann and C.~Rother, ``Learning less is more-6d camera localization via
  3d surface regression,'' 2018, pp. 4654--4662.

\bibitem{brachmann2019expert}
------, ``Expert sample consensus applied to camera re-localization,'' in
  \emph{ICCV}, 2019, pp. 7525--7534.

\bibitem{dsacstar}
------, ``Visual camera re-localization from rgb and rgb-d images using dsac,''
  2021.

\bibitem{arandjelovic2016netvlad}
R.~Arandjelovic \emph{et~al.}, ``Netvlad: Cnn architecture for weakly
  supervised place recognition,'' in \emph{CVPR}, 2016, pp. 5297--5307.

\bibitem{torii201524}
A.~Torii \emph{et~al.}, ``24/7 place recognition by view synthesis,'' in
  \emph{CVPR}, 2015, pp. 1808--1817.

\bibitem{zagoruyko2015learning}
S.~Zagoruyko and N.~Komodakis, ``Learning to compare image patches via
  convolutional neural networks,'' in \emph{CVPR}, 2015, pp. 4353--4361.

\bibitem{gclstm}
Y.~Seo \emph{et~al.}, ``Structured sequence modeling with graph convolutional
  recurrent networks,'' in \emph{International Conference on Neural Information
  Processing}.\hskip 1em plus 0.5em minus 0.4em\relax Springer, 2018, pp.
  362--373.

\bibitem{chebcomb}
M.~Defferrard \emph{et~al.}, ``Convolutional neural networks on graphs with
  fast localized spectral filtering,'' \emph{Advances in neural information
  processing systems}, vol.~29, 2016.

\bibitem{lee2021largescale}
D.~Lee \emph{et~al.}, ``Large-scale localization datasets in crowded indoor
  spaces,'' 2021.

\bibitem{taira2018inloc}
H.~Taira \emph{et~al.}, ``Inloc: Indoor visual localization with dense matching
  and view synthesis,'' in \emph{Proceedings of the IEEE Conference on Computer
  Vision and Pattern Recognition}, 2018, pp. 7199--7209.

\bibitem{resnet}
K.~He \emph{et~al.}, ``Deep residual learning for image recognition,'' in
  \emph{Proceedings of the IEEE conference on computer vision and pattern
  recognition}, 2016, pp. 770--778.

\bibitem{imagenet}
\url{https://www.image-net.org/}, (accessed Jan. 18, 2022).

\bibitem{4775883}
Z.~Wang and A.~C. Bovik, ``Mean squared error: Love it or leave it? a new look
  at signal fidelity measures,'' \emph{IEEE Signal Processing Magazine},
  vol.~26, no.~1, pp. 98--117, 2009.

\bibitem{wang2004image}
Z.~Wang \emph{et~al.}, ``Image quality assessment: from error visibility to
  structural similarity,'' \emph{IEEE transactions on image processing},
  vol.~13, no.~4, pp. 600--612, 2004.

\end{thebibliography}
\endgroup

\end{document}